# GLOBAL MINIMIZATION OF A QUDRATIC FUNCTIONAL : NEURAL NETWORK APPROACH

## L.B.Litinskii, B.M.Magomedov*

*Institute of Optical Neural Technologies Russian Academy of Sciences
119333, Moscow, Vavilov street 44/2, (095)-135-7802, e-mail: litin@iont.ru

The problem of finding out the global minimum of a multiextremal functional is discussed. One frequently faces with such a functional in various applications. We propose a procedure, which depends on the dimensionality of the problem polynomially. In our approach we use the eigenvalues and eigenvectors of the connection matrix.

## 1. Introduction

In various applications ([2], [3]) it is necessary to find out the global minimum of a quadratic form depending on $N$ binary variables $s_i = \{\pm 1\}$:

$$\min_{\vec{s}} \left\{ E(\vec{s}) = -(\mathbf{J}\vec{s}, \vec{s}) = -\sum_{i,j=1}^{N} J_{ij} s_i s_j \right\}. \quad (1)$$

$N$-dimensional vectors $\vec{s} = (s_1, s_2, ..., s_N)$ will be called *configuration vectors*. They define $2^N$ possible states of the system (the configurations), among which the optimal configuration with regard to *the objective function* $E(\vec{s})$ has to be found. The number of different states increases exponentially, when $N$ increases. Already if $N > 50$, it is almost impossible to solve the problem by means of direct enumeration.

In physical applications and in the theory of neural networks the matrix $\mathbf{J} = (J_{ij})_{i,j=1}^{N}$ is called *the connection matrix*. It is supposed to be a symmetric one. Then one can use the well-known dynamic procedure, where the objective function $E(\vec{s})$ behaves as the energy: it decreases monotonically during the evolution of the system. When started from an arbitrary configuration vector, the system rapidly gets into a local energy minimum. As a rule, the number of local minima is very large, and the problem is to start from "a good" initial state, which is located in the basin of attraction of the global minimum.

An algorithm of polynomial computational complexity allowing one to find out the global minimum of the functional (1) was proposed in [4], [5]. The algorithm approved itself for small $N \sim 20$. We generalized the approach on the functional of more general form, in particular, for nonsymmetrical connection matrices $\mathbf{J}$. In the next section we define the dynamic procedure and describe our approach of finding out the global minimum of the system. Then we present the numerical results. In the last section we estimate the computational complexity of the algorithm and its possible generalizations.

## 2. Minimization algorithm

Let $\vec{s}(t) = (s_1(t), s_2(t), ..., s_N(t))$ be the state of the system at the moment $t$. Let us define dynamics of the system by specifying the rule of evolution of coordinates:

$$s_i(t+1) = \text{sign}\left( \sum_{j=1}^{N} J_{ij} s_j(t) \right). \quad (2)$$

From the theory of neural networks and physical arguments ([1]-[3]) it is known that if

$$J_{ij} = J_{ji}, \quad J_{ii} = 0, \quad \forall i, j, \quad (3)$$

during the evolution of the system its energy $E(\vec{s})$ (1) decreases monotonically. In other words, starting from an arbitrary initial state, after several steps the system gets in the nearest local energy minimum.

We shall call the configuration vector $\vec{s}*$, which provides a global minimum for the functional (1), as *the ground state* of the system (the term comes from physics). The main idea of our approach is to search the ground state among those configuration vectors, which are the closest to the eigenvectors of the matrix **J** corresponding to the largest eigenvalues. We shall call such eigenvectors as "*the largest" eigenvectors*.

Indeed, the symmetric matrix **J** possesses the full set of the eigenvectors $\vec{\mathbf{f}}^{(i)}$. Let us number them according to the decrease of the corresponding eigenvalues $\lambda_i$:

$$\mathbf{J} \cdot \vec{\mathbf{f}}^{(i)} = \lambda_i \vec{\mathbf{f}}^{(i)}, \ \lambda_1 \geq \lambda_2 \ldots \geq \lambda_N, \ (\vec{\mathbf{f}}^{(i)}, \vec{\mathbf{f}}^{(j)}) = \delta_{ij}.$$

Since the diagonal elements of the matrix are equal to zero, a part of the eigenvalues is positive, and another part is negative. Now the functional (1) can be expressed in terms of the eigenvalues and the eigenvectors of the matrix **J**:

$$E(\vec{\mathbf{s}}) = -\left(\lambda_1 (\vec{\mathbf{s}}, \vec{\mathbf{f}}^{(1)})^2 + \ldots + \lambda_N (\vec{\mathbf{s}}, \vec{\mathbf{f}}^{(N)})^2\right). \quad (4)$$

The expression (4), first, allows one to find the lower estimate for the global minimum of the functional (1): $E(\vec{\mathbf{s}}^*) \geq -\lambda_1 N$. Second, from Eq.(4) it is clear that one has to look for the solution of the problem (1) among those configuration vectors $\vec{\mathbf{s}}$, which have the smallest possible projection onto the eigenvectors corresponding to the negative eigenvalues. Indeed, with accuracy to the sign $E(\vec{\mathbf{s}})$ is a weighted sum of the squared projections of the configuration vector $\vec{\mathbf{s}}$ onto eigenvectors $\vec{\mathbf{f}}^{(i)}$. The weights entering the sum (4) are $\lambda_i$. Then, the larger the projection of a configuration vector $\vec{\mathbf{s}}$ onto the subspace spanned over "the largest" eigenvectors, the smaller is the value of the functional $E(\vec{\mathbf{s}})$.

To clarify the aforesaid, we analyze the example when the eigenvalue $\lambda_1$ exceeds essentially all other eigenvalues:

$$\lambda_1 \gg \lambda_i, \ i = 2, \ldots N. \quad (5)$$

(One faces such a situation rather often.) Then in the expression (4) we can restrict ourselves with the first term only:

$$E(\vec{\mathbf{s}}) \approx -\lambda_1 (\vec{\mathbf{s}}, \vec{\mathbf{f}}^{(1)})^2. \quad (6)$$

Evidently, here the ground state is the configuration vector the closest to $\vec{\mathbf{f}}^{(1)}$. It is very simple to find out this configuration vector: its coordinates are defined by the sings of the coordinates of $\vec{\mathbf{f}}^{(1)}$:

$$s_i^* = \text{sign}(f_i^{(1)}), i = 1, \ldots, N \Rightarrow$$
$$\Rightarrow (\vec{\mathbf{s}}^*, \vec{\mathbf{f}}^{(1)}) = \sum_{i=1}^N | f_i^{(1)} | \geq |(\vec{\mathbf{s}}, \vec{\mathbf{f}}^{(1)})| \ \forall \ \vec{\mathbf{s}}.$$

Thus, if the maximal eigenvalue $\lambda_1$ of the matrix **J** exceeds all the other eigenvalues essentially, the solution of the problem (1) is very simple. Complications begin when there are some eigenvalues comparable with the maximal one. In this situation in the functional (1) the contributions from the aforementioned eigenvalues compete with each other. It can occur that the global minimum is achieved on the configuration vector, which is the nearest not to the first eigenvector, but to some other eigenvector. However, from the expression (4) it is evident that the global minimum cannot be achieved on the configuration vector orthogonal to the subspace spanned over "the largest" eigenvectors. This argument defines the main idea of our approach: to find out the ground state of the functional (1) the dynamic system (2) has to start from the configuration vectors, which are close to "the largest" eigenvectors.

### 3. Computer simulation

<u>Matrices of small dimensions.</u> We generated 1500 symmetric random matrices **J**: in groups of 500 matrices of dimensionality (15x15), (16x16) and (17x17), respectively. Off-diagonal matrix elements were chosen randomly and independently from the interval [-4,+4], and all the diagonal elements were equal to zero.

For each of the matrices we found out the global minimum of the functional (1) with the aid of the direct enumeration. Simultaneously for a given matrix we calculated its eigenvalues and eigenvectors. Then for each eigenvector $\vec{\mathbf{f}}^{(i)}$ with a positive eigenvalue we found 3 the most close configuration vectors. As a result we had a set of approximately 20 configuration vectors, which were used as the start states for the dynamic system (2). We determined the local minimum in which our system got.

It turned out, that if the dynamic system was started from one of the 3 configuration vectors the closest to the largest one, the probability (averaged over the ensemble of all 1500 random tests) to get into the global minimum was equal to 0.8. If the dynamic system was started from 20 chosen configuration vectors, it got to the global minimum with the probability equal to 0.97.

In other words, for such values of $N$ the basin of attraction of the global minimum is reliably "covered" by the set of 20 configuration vectors, which are the closest to "the positive" eigenvectors of the matrix **J**.

<u>Matrices of large dimensions.</u> One of the problems arising for large $N \sim 10^2 - 10^3$ is that because of the dimensionality of the problem, it is impossible to find the ground state of the system. In this case a reasonable check up of our algorithm is to compare the largest depth of the local minimum obtained with the aid of our algorithm, with the largest depth of the local minimum determined by the random search.

The calculation complexity of our method is defined by the time required for calculation of the eigenvalues and eigenvectors of the connection matrix. In order of magnitude this time is equal to $N^3$. Since for the system started from an arbitrary initial configuration, about $N^2$ operations are needed to get into the nearest local minimum, the comparison between the two methods will be correct, if the system is started from $N$ random configurations. In this case the calculation complexities for the both approaches are the same.

For different $N$ we performed a set of calculations involving 1000 random matrices. The first type of examined matrices was analogous to the aforementioned one. We used random symmetric matrices with $N = 100$, 200, 300, 400, 500 and 750. In physics the same kind of matrices is used when describing the so called *spin glass* characterized by multiply degenerated ground state. For this system the number of local minima is exponentially large. All these minima have approximately the same depth, distributed with a small dispersion around some mean value. When $N$ increases, the dispersion of the distribution tends to zero. In other words, here a pronounced global minimum is absent, and the search of the lowest local minimum is especially difficult.

The results of this experiments are very interesting. At the moment they are under processing. Their detailed analysis will be given elsewhere. Here we present the first obtained results only.

For $N = 100$, 200, 300, 400, 500 and 750 the probability (averaged over 1000 random tests) that the lowest local minimum obtained with the aid of our method, is deeper than the local minimum calculated by means of the random search is equal to 0.37, 0.56, 0.57, 0.61, 0.62 and 0.75, respectively. Thus, it can be stated that if the system is started from the configuration vectors the closest to the eigenvectors, more than in the half of the cases we obtain the deeper local minimum than with the aid of the random search.

We would like to note, that the probability of the success of our method increases, if $N$ increases. Next, when $N$ increases, the difference between the value of lowest local minimum obtained with the aid of our method, and the one found by means of the random search increases too. This result has to be defined more exactly.

The second kind of the examined matrices is the Hebb connection matrices, which are well-known in the theory of associative memory. In fact, they are the correlation matrices of $p$ random $N$-dimensional binary vectors (so called *the patterns*). It is known that for $N >> 1$ and $p/N > 0.14$ the energy landscape of the system resembles the energy landscape of spin glass: it has an exponentially large number of local minima, and it has no a pronounced global minimum. If $p/N \sim 0.05$ or less, the system has $p$ global minima coinciding with the patterns, and an exponentially large number of local minima. In this case there is a gap between the energies of local minima and the energies of global minima.

For the Hebb matrices to the moment we have done the calculations with $p/N = 0.02$ only. The global minimum of the functional can be reliably found if the system is started from $p$ configuration vectors which are the closest to the $p$ largest eigenvectors. An unexpected result is, that the system gets to its global minimum more often when started from the configuration vector which is the closest to the second of the largest eigenvectors, and not to the first one. The examination of the underlying reasons is in progress.

## 4. Conclusions

Our approach suggests that a deep local minimum of the functional (1) can be found during the time of the order of $O(N^3)$. There are no restrictions on the matrix $\mathbf{J}$, except it has to be a symmetric one.

This restriction can also be discarded. Indeed, let the matrix $\mathbf{J}$, entering Eq. (1), be a nonsymmetric one. The discussed approach can not be directly applied in this case, because such a matrix does not possess a full set of eigenvectors. However, the matrix $\mathbf{J}$ can be symmetrized with the aid of the transformation $\tilde{\mathbf{J}} = (\mathbf{J} + \mathbf{J}^\mathbf{T})/2$, where $\mathbf{J}^\mathbf{T}$ is the transpose of the matrix $\mathbf{J}$. It is easy to check, that for any configuration vector the values of the quadratic functional (1) are the same for the matrices $\tilde{\mathbf{J}}$ and $\mathbf{J}$. Then our approach can be applied to the symmetric matrix $\tilde{\mathbf{J}}$. Note, the requirement that the diagonal elements in (3) are equal to zero is not a restriction, since the local minima of the functional (1) are independent from these diagonal elements (this can be easily checked by direct calculation).

Usually, the functional that has to be minimized, contains not only quadratic, but also a linear term with respect to binary variables. Such a problem can be easily reduced to the basic form of the functional (1) by introducing a new fictitious coordinate. Then, an additional column and an additional row appear in the connection matrix.

It would not be possible to complete this work without stimulative support of our colleague B.V.Kryzhanovsky.

The work was supported by the program "Intellectual Computer Systems" (the project 2.45) and in part by Russian Basic Research Foundation (grant 03-01-00355) and project 1.8 of DITCS RAS.